# Human Motion Intent Inferencing in Teleoperation Through a SINDy Paradigm

Michael Bowman*, and Xiaoli Zhang, *Member, IEEE*

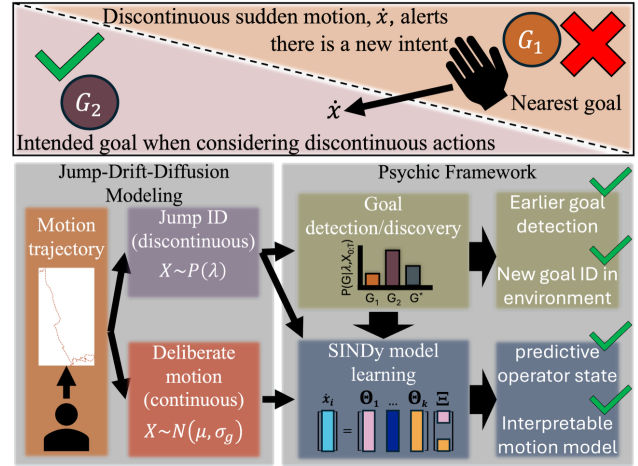

Fig. 1: Intent inference methods in teleoperation often ignore subtle motion that can be strong indicators for a sudden change in intent. We model these small indicative motions through a jump-drift-diffusion stochastic differential equation, and apply a SINDy model on approximate coefficients with goal approximation as a control input. Our method, called Psychic, allows for goal intent inference, and operator predictive states with interpretable motion models unlike traditional regression methods like negative log-likelihood estimates.

*Abstract*—Intent inferencing in teleoperation has been instrumental in aligning operator goals and coordinating actions with robotic partners. However, current intent inference methods often ignore subtle motion that can be strong indicators for a sudden change in intent. Specifically, we aim to tackle 1) if we can detect sudden jumps in operator trajectories, 2) how to appropriately use these sudden jump motions to infer an operator's goal state, and 3) how to incorporate these discontinuous and continuous dynamics to infer operator motion. Our framework, called Psychic, models these small indicative motions through a jump-drift-diffusion stochastic differential equation to cover discontinuous and continuous dynamics. Kramers-Moyal (KM) coefficients allow us to detect jumps with a trajectory which we pair with a statistical outlier detection algorithm to nominate goal transitions. Through identifying jumps, we can perform early detection of existing goals and discover undefined goals in unstructured scenarios. Our framework then applies a Sparse Identification of Nonlinear Dynamics (SINDy) model using KM coefficients with the goal transitions as a control input to infer an operator's motion behavior in unstructured scenarios. We demonstrate Psychic can produce probabilistic reachability sets and compare our strategy to a negative log-likelihood model fit. We perform a retrospective study on 600 operator trajectories in a hands-free teleoperation task to evaluate the efficacy of our opensource package, Psychic, in both offline and online learning.

## I. INTRODUCTION

### A. Intent Inferencing in Teleoperation and Background

Teleoperation is often reliant on developing intent inference models to understand operator's actions and goals to alleviate cognitive burden and workload. Intent inferencing often relies on behavioral cues to identify actions and goals to spare operator's from explicitly stating their goals [1, 2]. In teleoperation, predicting an operator's intent is the first step to provide assistance in shared control as it aligns a robotic partner to the operator by understanding what high-level goal an operator is trying to accomplish and why they are trying to complete this goal.

However, inferring an operator's intent in unstructured teleoperation environments is challenging [3, 4]. This challenge is due to the operator's inconsistent motions, lingering in a single pose, erratic suboptimal behaviors, and transitions between goals. Current work in intent inference investigates grasping goals or analyzing single short-horizon actions [5-8]. Models have attempted to infer intent through data-driven means such as Neural Networks [8], Support Vector Machines, Hidden-Markov models [9-11] and Deep Reinforcement Learning (DRL) models [3, 12, 13]. When discrete goal locations are in play, probabilistic methods often shortcut the inference via an inverse distance method to the nearest goals to disambiguate an operator intended goal state (seen in Fig. 1). In goal discovery settings, maximum entropy methods over belief states are a common strategy [14, 15]. However, determining the emergence of a new goal state is still an open problem within the intent inferencing field. For instance, it is unclear and difficult to determine if an operator makes large motions to self-reset an attempt to approach an object in teleoperation settings. Further challenges emerge when there are unmodeled obstacles and goals in the environment. Alternatively, some research forgoes modeling goals and opts to infer *how* an operator is moving directly.

Predicting motion has long been studied where a common operator motion dynamics model is the minimum jerk principle [16]. Specifically, in reaching tasks it has been assumed users move in Conic trajectories in decoupled dimensions [17]. These analytical models work for large behaviors but have not been explored when operators try to refine subtle motions. Analytical modeling applies when an operator truly has a strong understanding of their dynamics and move with decisive actions while not in control of a robot. In teleoperation tasks, operators move more unnatural due to disembodiment problem with a robot [18, 19]. This leads to them deploying strategies that differ from natural intuitive motion [20]. Others have formulated operators motion with a drift-diffusion model to conform with Markovian

*This work is funded by NSF under grant 2114464.

Michael Bowman is a postdoctoral fellow in the Cancer Biology Department at University of Pennsylvania, Philadelphia, PA 19104 (corresponding author*: Michael.Bowman@pennmedicine.upenn.edu)
Xiaoli Zhang is an Associate Professor in the Mechanical Engineering Department at Colorado School of Mines, Golden CO, 80401. (xlzhang@mines.edu

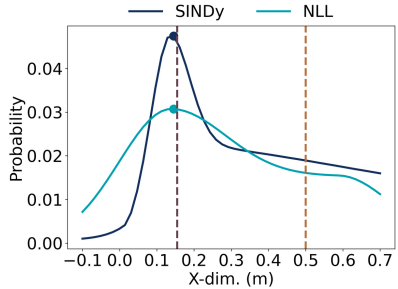

Fig. 2: Probability Density Functions (PDF) for an NLL and SINDy derived KM terms to the jump-drift-diffusion equation. Dashed lines represent goals. Points represent the MAP value of the PDFs.

assumptions for DRL strategies [21, 22]. These strategies require human demonstrations, and have been explored to account for similarity of trajectories [23]. While these approaches work for smooth motions they do not account for behaviors that come with discontinuous motion including sudden starts or hesitations. When an operator moves unexpectedly and causes confusion, these predictive models hold lower confidence in their prediction. We propose to account for the discontinuous motions in our model by expanding the drift-diffusion model to a jump-drift-diffusion model. We rely on a structured machine learning technique that infuses analytical modeling with a data-driven approach to recover the dynamics of the jump-drift-diffusion model.

*B. Sparse Identification of Nonlinear Dynamics*

A new machine learning paradigm called Sparse Identification of Nonlinear Dynamics (SINDy) has been developed for system identification [24]. The system identification field has produced state-space inference methods such as autoregressive models (NARMAX) [25] and recent developments in DRL have been paired with Model Predictive Control (MPC) [24, 26]. A benefit of SINDy over these other strategies is it has high interpretability and reduces overfitting of specific curated datasets. Specifically, it has high utility in approximating nonlinear systems that contain foundational analytical forms for system dynamics, even in chaotic systems. For instance, if a hypothesized functional form of the dynamics is given, SINDy can recover coefficients and terms that best explain the data for a particular perturbation. SINDy has not been used to identify an operator's motion dynamics nor to infer intent in teleoperation. To the authors' knowledge, the closest related work is the use of Koopman operators for predicting the system dynamics of a lunar lander game and applying a model-based shared control approach with the fitted model [27]. This work shows promise for a carefully curated basis function, and features to be used in model-based control. While our paper focuses on intent inference, this previous work lays down a roadmap for applicability downstream. We cover the foundational aspects of using SINDy for intent inference in II.B.

*C. Contributions*

A retrospective analysis of human motion gathered from studies in teleoperation culminating in 600 user trajectories [28-30] is conducted. This paper's aim is to tackle the open problems of detecting existing goals and discovering new ones on the fly. The second aim is to probabilistically predict operator motions. Our contributions are 4-fold:
1. Development of a novel stochastic, interpretable motion model, via a jump-drift-diffusion process, that describes operator motion which we infer through a SINDy regression.
2. Development of a novel use of the jump-drift-diffusion model with a statistical outlier detection algorithm to identify and discover goals.
3. Deployment of a probabilistic reachability set for the inferred operator's model-based motion.
4. Demonstration of the feasibility of the method on both offline and online modeling in two separate teleoperation scenarios.

## II. METHODS

The overall intent inference system is summarized in Fig. 1, where user observations are collected and a stochastic differential equation to model their dynamics is applied. A traditional regression strategy to solve this equation is through a negative log-likelihood (NLL) estimate which will serve as a baseline. However, a SINDy algorithm is used to infer the functional form of different properties of the stochastic equation (i.e, an operator's drift behaves squared or cubic rather than linear). This is a necessary requisite for model-based optimal control strategies including MPC or DRL which is left for future endeavors. Further, the framework identifies potential goals, changes in goals, and reachability sets during trajectories. Our open-source codebase is called Psychic (found at https://github.com/namwob44/Psychic). We will use an actual operator trajectory to illustrate the methods throughout.

*A. Jump-Drift-Diffusion Modeling with Known Goals*

We extend Markovian motion dynamics for humans from drift-diffusion modeling to include discontinuous motions through jump-processes. First, it is assumed each cartesian motion is independent to one another, and a parallel formulation can be deployed. For a state trajectory vector of a single dimension, $X$, at starting time, $s$, and a terminal time $t$, the aim is to model the transition from $X_s$ to $X_t$ toward goal states, $g \in G$, through a drift-diffusion model. This follows a stochastic differential equation of the form:

$$dX_t = \mu_g(X_s, g)dt + \sigma_g dW_t \quad (1)$$

Effectively, the drift term, $\mu_g(X_s, g)$, acts as a mean reversion (or an attractor) toward a goal state commonly seen in econometrics. The $\sigma_g$ is the diffusion term representing the magnitude on a continuous Weiner noise process, $dW_t$. We extend (1) by adding an additional Poisson jump term which accounts for discontinuous actions (sudden starting and stopping motions) of an operators.

$$dX_t = \mu_g(X_s, g)dt + \sigma_g dW_t + \beta_g dP_t \quad (2)$$

$\beta_g$ is an encompassing term for parameters in a Poisson process $dP_t$, including a jump rate ($\lambda$), intensity ($\mu_\beta$), and variance ($\sigma_\beta^2$). The above stochastic differential equation has several unknown parameters that must be approximated given a full state trajectory for a given goal. One could solve for these parameters using an NLL optimization to find a best fit for each unknown. This strategy serves as a baseline approach in the experimental setup. The probability density function for state transitions toward $X_t$ given $X_s$ and $g$ is shown in (3). This equation follows a combination of a Poisson and Normal distributions with an infinite series:

$$P(X_t|X_s, g) = \sum_{\kappa=0}^{\infty} \frac{\lambda^\kappa \tau^\kappa e^{-\lambda \tau}}{(\kappa!)\sqrt{2\pi(\sigma_g^2 \tau + \kappa \sigma_\beta^2)}} \exp\left(-\frac{(X_t - X_s - \mu_g \tau - \kappa \mu_\beta)^2}{2(\sigma_g^2 \tau + \kappa \sigma_\beta^2)}\right) \quad (3)$$

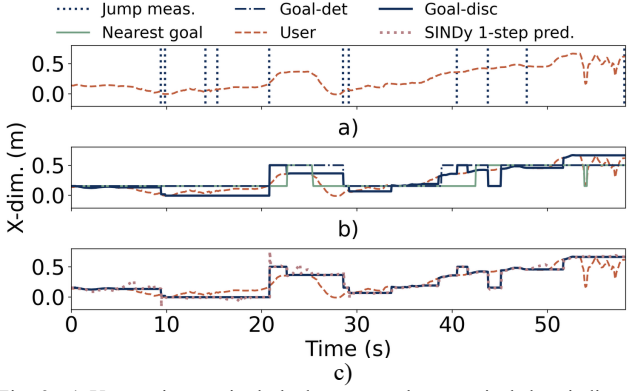

Fig. 3: a) User trajectory is dashed orange, where vertical dots indicate detected jumps occurred using equation (8). b) on top of the user trajectory we overlay, teal is the nearest goal, while the others are the Psychic inferred goals. Detection mode is dashed blue which stabilizes better than discover mode (solid blueline). c) The light brown is predicting 1-step for goal-disc, which aggressively changes points at jumps.

After models are fit for each $g$, Expectation Maximization can be used to obtain posterior probabilities and classifications for the operator's most likely goal. Further, by taking the maximum-a-posteriori (MAP) of the distribution, predictive actions across the state space can be made. An example of the PDF for a two-goal system is shown in Fig. 2. While this model fit is possible with a full trajectory, it is difficult to rely on the parameter fits for online strategies needed in teleoperation.

Therefore, alternative formulations using Kramers-Moyal (KM) coefficients are developed to reasonably determine the properties of the jump-drift-diffusion modeling by following [31]. Even number coefficients relate distribution properties including variance, kurtosis, and hyper-kurtosis, which are the contributing factors for diffusion and jump properties. These properties can be summarized by taking the moments of the stochastic distribution in (2):

$$M^{(n)}(x, t) = \lim_{dt \to 0} \frac{1}{dt} \langle |x(t + dt) - x(t)|^n \rangle \quad (4)$$

where the first moment, $M^{(1)}$, acts as the drift coefficient ($\mu_g$) and is the primary driver for the motion toward a goal.

$$M^{(1)}(x, t) = \frac{dx}{dt} \quad (5)$$

Taking the second moment, $M^{(2)}$, can be approximated as the following equation and relates to the variance of the trajectory motion containing both diffusive and jump variances ($\sigma_g^2, \sigma_\beta^2$).

$$M^{(2)}(x, t) = \frac{dx^2}{dt} \quad (6)$$

The $M^{(4)}$ and $M^{(6)}$ are higher order terms needed to recover jump processes by cutting through the more dominant diffusion features contributing to discontinuous processes.

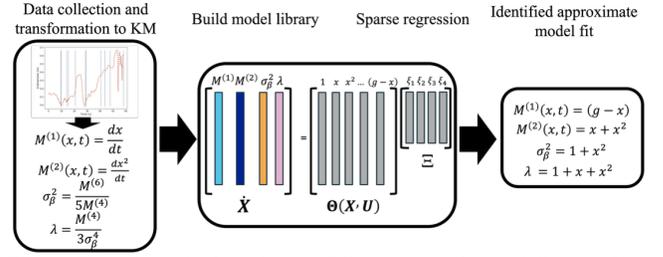

Fig. 4: The schematic for using a SINDy strategy for modeling the KM terms, goal states, and an example output it produces.

$$M^{(4)}(x, t) = \frac{dx^4}{dt}, M^{(6)}(x, t) = \frac{dx^6}{dt} \quad (7)$$

These play explicit roles in recovering $\sigma_\beta^2$ and $\lambda$ which can be recovered by relaxing the assumption of $\mu_\beta = 0$.

$$\sigma_\beta^2 = \frac{M^{(6)}}{5M^{(4)}}, \lambda = \frac{M^{(4)}}{3\sigma_\beta^4} \quad (8)$$

The KM formulation requires 6th order data to ensure we model jump processes. Modeling high order data is not unfamiliar for modeling human motion (minimum jerk models require 5th order data to approximate motion). The work presented in [31] contains approaches to determine whether a purely diffusive or a jump-diffusion process better reflects the dynamics an operator is exhibiting. The focus of this work is not to identify operators based on their diffusive or jumpy behaviors; therefore, we opt to identify purely diffusive behaviors through the following approximation.

$$\sigma_k^2 = M^{(2)} - \lambda * \sigma_\beta^2 \quad (9)$$

The benefit of this approximation is if jumps are inconsequential then $\lambda \to 0$, thereby having a strictly diffusive model for the operator behavior. Fig. 3 shows an example of jumps being identified within a trajectory. A major benefit of KM coefficients is it allows for the dynamical processes to be modeled independently. However, explicitly account for $g$ in the formulation and account for functional form of the KM coefficients is still required.

*B. Providing Jump-Drift-Diffusion Terms by SINDy*

Next, it is necessary to identify the functional form of KM coefficients throughout a trajectory to recover each component of the dynamics model in terms of $X$ and $g$. We acknowledge stochastic differential equations have already applied SINDy to other applications [32, 33]. SINDy methods begin with the formulation of a differential equation $\frac{d}{dt}X = f(X, U)$ where $U$ is a matrix of control signals (in this case $g$). Then a library of nonlinear candidate basis functions $\Theta(X, U)$ and a vector of parameter coefficients $\Xi = [\xi_1, \xi_2, \dots, \xi_n]$ to approximate the sparse representation of $f(X, U)$ are defined. A Stepwise Sparse Regressor (SSR) [32, 34, 35] is then used:

$$\dot{X} = \Theta(X, U)\Xi \quad (10)$$

Each KM coefficient is treated as an independent $\dot{X}$ with an independent $\Theta(X, U)$ to identify a sparse library to approximate the parameters, $\mu_g, \sigma_g^2, \lambda$, and $\sigma_\beta^2$. An example of the schematic for SINDy is shown in Fig. 4.

A benefit of this strategy is the transformation of the goal into continuous space, which no longer necessitates model fits for discrete goals like the NLL models. Once $\Xi$ are found, we can predict components of the dynamical equation in (2). However, operator actions are still stochastic, so we opt to predict their next actions using probabilistic inference in (3) with the SINDy approximated KM coefficients, with the best choice being the MAP.

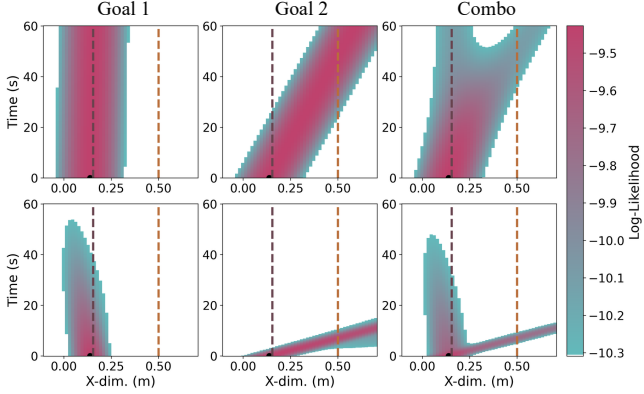

Fig. 5: An example of the log-likelihood reachability sets for a single dimension across a future time horizon (y-axis). Two goals are shown with dashed lines. The 1st column is the reachability set for only going toward goal 1, and the 2nd column is the reachability set toward goal 2. The last column is combination of the goal specific reachability sets. The top row is NLL estimate and the bottom row is the SINDy estimate. Each slice in time shows the likelihood that the operator will be at that position at that specific time in the future. This is based on the drift, jump, and diffusion terms.

*C. Incorporating SINDy for Goal Detection and Discovery*

This work proposes SINDy can be used for unknown goal inference. In doing so, goals must be transformed from static states into dynamic ones that evolve over time. First, a functional form of a dynamic goal equation is established and a candidate library terms $\Theta(g, U)$, is defined. We draw inspiration of another intent inference modality, eye gaze. In eye-gaze, concepts of dwell time and fixations determine a goal for hand motion through statistical measures [1]. If no deliberate motion occurs, the operator has settled on a new goal (fixation). Settling can be viewed as dispersion/concentration of deliberate motion through the coefficient of dispersion:

$$R = \frac{\mu_{M(1)}}{\sigma^2_{M(1)}} \quad (11)$$

Through window techniques and robust measures, the median, $\bar{R}$, and median absolute deviation (MAD) are obtained. Afterwards, the observed concentration is scored, $S_{settled}$, by ranking it through a standard normal error CDF.

$$S_{settled} = \frac{1}{2}\left(1 + \text{erf}\left(\frac{\bar{R}-R}{\sqrt{2}\,MAD(R)}\right)\right) \quad (12)$$

To determine if an action is moving, the compliment of the $S_{settled}$ is taken with $S_{move} = 1 - S_{settled}$. While finding a settled goal is one aspect of interest, there is a competing scoring function to identify goal switching based on jump information $(\lambda, \sigma^2_\beta)$. Jumps indicate a sudden change in operator motion which we interpret as change in intent. A method called Empirical CDF Outlier Detection (ECOD) is used to identify extreme outliers in jump events [36]. For brevity we omit the derivation, but assign the $S_{jump}$ to be 1 if a point is an outlier, and 0 otherwise. The assumption is jumps are induced purposefully in a desired direction. After, $S_{jump}$ is multiplied with the operator's measured motion direction, $v_{Dir}$, and with a magnitude from the estimated $\sigma_\beta$. The same concept applies to drift motion, $S_{move}$.

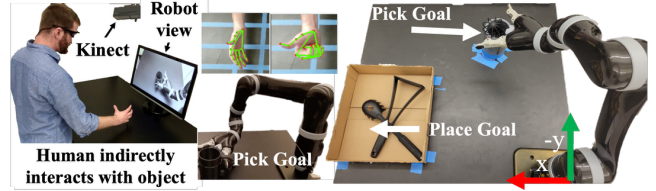

Fig. 6: The studies conducted hands-free teleoperation, where operators could freely move around and an overhead camera collected hand information. We requested recorded operator motion and known goal locations for each scenario. Scenario 1 had a single grasping point (central robot photo), while scenario 2 contained two goals including a pick and place goal (right robot photo).

When $S_{settled}$ is high, an operator is likely not moving, implying a new goal has emerged. When a new goal is found, $\dot{g} \approx 0$, which we interpret as $g \approx X$. Lastly, the formulation for goal derived motion accounts for known goals in an environment by using a softmax proximity weighting scheme ($w_k$) for the current predicted goal toward known ones ($g_k$). These combined aspects describe $\dot{g} = f(g, X, U)$ dynamics as follows:

$$\dot{g} = \xi_1 S_{settled}(X - g) + \xi_2(S_{move})v_{Dir} + \xi_3 S_{jump}\sigma_\beta v_{Dir} + \sum_k^K \xi_{3+k}(S_{move})w_k(g_k - g) \quad (13)$$

The $\xi$'s in front of each term are nominally weighting terms, in this instance with equal contribution. Discovering a moving goal may not always be desired, therefore two separate modes are made: 1) detection mode (goal-det) which snaps to the nearest known goal, and 2) discovery mode (goal-disc) which continually looks for a settled goal. The running example in Fig. 3 shows operator motion, nearest neighbor goal, goal-det, and goal-disc. Alternatively, SINDy can estimate each component's contribution ($\xi$'s) in the goal inference, with each term making up the library, $\Theta(g, U)$. The running example in Fig. 3c shows operator motion, goal-disc, and SINDy 1-step prediction on goal-disc (equation (13)).

*D. Producing Reachability Sets*

Reachability sets identify potential state bounds an operator can reach at a given time [37]. Rather, than compute the full forward reachability sets, a probabilistic approach to determine a suitable substitute by using (3) and spanning discretized possible states for forward predictions is presented. The goal space ($g_k \in G$) is marginalized for each motion dimension ($n \in N$). We assume independence in each spatial dimension and multiply the likelihoods together. Afterwards, the maximum likelihood for each voxel across the temporal axis is found.

$$P(X|X_s) = \max_t \log \prod_{n=0}^N \sum_k^K P_n(X_t|X_s, g_k) \quad \forall_X \quad (14)$$

In (14), time is viewed as a hyperparameter for how far in advance the prediction is carried out. Figure 5 shows an example of reachability sets without the temporal axis maximizer for the running example with the point starting at time=0 and the prediction horizon being 60s with the NLL and SINDy in detection mode for the 2 known goals. By taking slices at time intervals, prediction horizons are effectively created for each goal where more examples are in the results section of III.C.

## III. EXPERIMENT SETUP AND RESULTS

### A. Experimental Setup

The retrospective study uses data from three studies [28-30]. Operator motion was collected from a hands-free teleoperation task, shown in Fig. 6, for two scenarios: 1) a single goal reaching task, and 2) a two-goal pick and place task. In total 360 trajectories for the single goal and 240 trajectories for the two-goal scenarios were analyzed from 25 operators across the 3 studies. Operators' motion was collected via hand tracking from MediaPipe at 20Hz. The trajectories XYZ positions will be considered in the statistical analysis, however, for clarity and readability only 1D and 2D motions will be used as visual examples. Online methods performed trajectory playback, so model retraining occurred at each new time step including on data up to time $t$.

There are two separate analyses, the first is investigating goal detection and discovery, while the second is investigating performance of SINDy for model-based inference. The first analysis uses nearest neighbors (NN) as a baseline for goal detection which is a common shortcut to goal inference via Voronoi cells [38]. NN is compared to the goal-det and goal-disc modes. For the second analysis, the NLL is compared to the SINDy model fit for the three separate goal modes (NN, goal-det, goal-disc). These 6 models are compared in both online and offline modes.

### B. Goal Detection and Discovery Analysis

We hypothesize measured jumps indicate an induction of goal/intent switching by operators. For each trajectory we quantify the geometric means and 95% confidence intervals for the number of jumps identified, and number of times goals switched using NN, goal-det (shown in Fig. 3b as an example), or the more continuous goal-disc. A 2-way ANOVA is performed on the measures between the Task Scenarios (F=154.78, p=1.78E-34) and Goal Strategies (F=314.48, p=4.2E-172). The interaction between both task and strategies (F = 98.368, p = 4.87E-60) shows there is a dependence on both measures. As expected, more jumps and goal switches are in a multi-goal task and the goal-disc method continuously adapts (implying goal-disc continually follows operator actions to new settled locations as stated in the II.C with $g \approx X$).

TABLE I. KNOWN GOAL INFERENCE SUMMARY STATISTICS

| Goal Strategy | Task Scenario | Jumps Detected | Number of Goal Switches |
|---|---|---|---|
| NN | Single Goal | 10.46[9.96, 10.98] | None |
| | Two Goal | 16.59 [14.59, 18.88] | 11.81[10.52,13.25] |
| goal-det | Single Goal | 10.46[9.96, 10.98] | None |
| | Two Goal | 16.59[14.59, 18.88] | 28.84[25.57, 32.53] |
| goal-disc | Single Goal | 10.46[9.96, 10.98] | 209.92[192.99, 224.00] |
| | Two Goal | 16.59[14.59, 18.88] | 474.36[416.14, 540.72] |

Afterwards, we investigate how effective SINDy can be used to predict the dynamic goal change. An offline and online SINDy model is used to compute the 1-step forward prediction and is compared to the measured goal trajectory it is trying to follow. A Residual Sum of Squares (RSS) is used where the geometric mean and 95% confidence interval is found in Table II. A 2-way ANOVA between Prediction Strategies and Task Scenarios is conducted separately for offline and online models using the RSS values. For the offline model there is a significant difference between the tasks (F=64.66, p=2.11E-15), and no significance between the strategies (F=0.027, p=0.869), nor between their interaction (F=0.102, p=0.749). This is interpreted as the SINDy model holding up for both prediction strategies. However, as seen in Fig. 3, the 1-step prediction tends to overshoot goals when jumps are detected and switching goals are involved. For the online model there is no statistical significance for strategies (F=0.0007, p=0.97), tasks (F=1.5, p=0.22), nor their interaction (F=0.001, p=0.97). While this informs goal-det and goal-disc may perform the same, the geometric mean and 95% confidence intervals have a switch in error rates between tasks. Possible reasons for this discrepancy are 1) models are locked into a local minimum early on, or 2) due to the ECOD strategy struggling to identify outlier jumps early on in shorter trajectories (single goal task). While Table II shows more nuance and exploration is needed to the mechanisms of goal inference for online learning, the notions gained from the offline learning shows more reasonable promise.

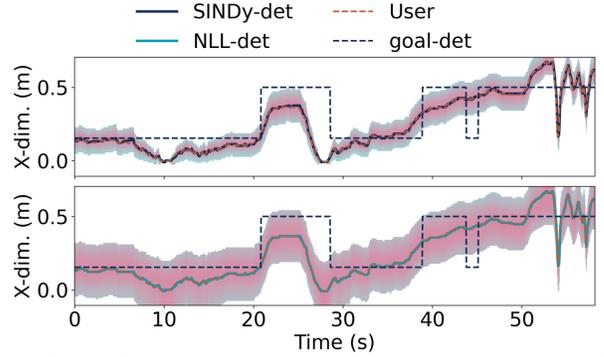

Fig. 7: Plots are 1-step prediction with single timestep reachability sets. The top row is the SINDy-det and the bottom is the NLL-det. We see both methods perform well at predicting operator motion (dashed orange), with SINDy having much tighter bounds than NLL. The MAP predictors are nearly indistinguishable from user input at 1-step predictions for both methods, where bounds are more informative for state actions.

TABLE II. SINDY 1-STEP PREDICTION MODEL SUMMARY STATISTICS

| Prediction Strategy | Task Scenario | Offline RSS (m²) | Online RSS (m²) |
|---|---|---|---|
| goal-det | Single Goal | 0.05 [0.03,0.07] | 1.27 [1.07, 1.50] |
| | Two Goal | 3.73 [3.01,4.61] | 0.22 [0.17, 0.29] |
| goal-disc | Single Goal | 0.25 [0.23,0.27] | 2.40 [2.17, 2.66] |
| | Two Goal | 3.38 [2.8,4.09] | 0.23 [ 0.18, 0.29] |

### C. Model Comparison

#### 1) Qualitative Analysis Reachability Sets

The differences between the detection modes, NLL-det and SINDy-det, are qualitatively evaluated. Figure 7 contains reachability sets for a single timestep prediction horizon along the running example for the 1-D trajectory with the two-goal scenario for NLL-det and SINDy-det. Both models chose the next state based on MAP for each prediction horizon timestep using (3) after their respective model fit. The actions chosen mirror the operator well, however, bounds for NLL-det are more conservative than the SINDy-det demonstrating less certainty in actions. SINDy-det holds tighter bounds as the dynamic approximation is more certain.

Further we show how the reachability sets can be sliced at various time points in the 2D scenario for the single goal task

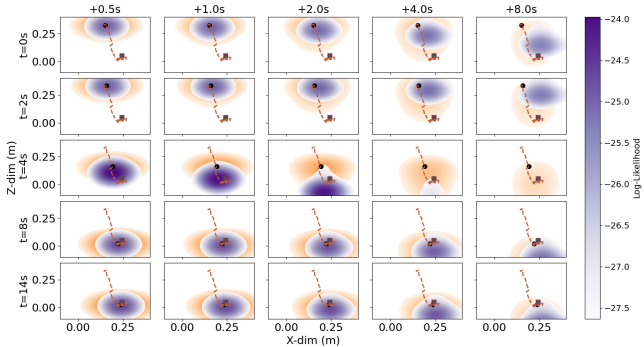

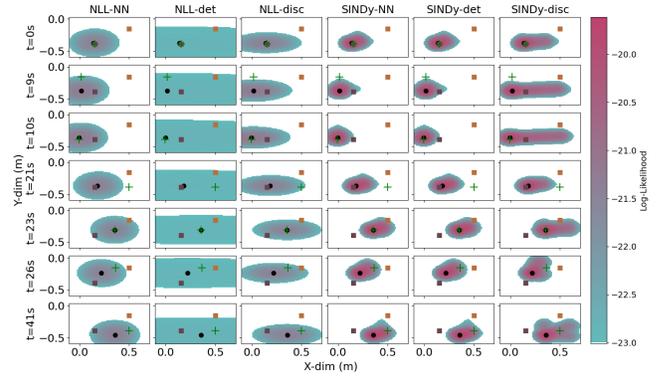

Fig. 8: Predictive slices of reachability sets for NLL-det (orange) and SINDy-det (purple) overlayed for a single goal (square on plots) scenario in the XZ plane. Columns represent the prediction horizon slice while rows represent distinct time points along the trajectory (orange dashed line and black circle for current location). This trajectory's entire time was only 14s, and a 1s prediction horizon appears most reasonable. We see the SINDy-det model is more aggressive in the forward predictions assuming operators overshoot goals than NLL-det which is more conservative

in the XZ plane. Figure 8 contains an example of prediction horizons (t = [0.5, 1.0, 2.0, 4.0, 8.0]) for NLL-det and SINDy-det overlayed on one another. Doubling times are chosen as the different prediction horizons (columns) to obtain a better sense of the relative differences for lookahead functions. The true full length trajectory is shown (dashed orange) and is only 14s long in each plot. Each row is the current state (black dot) at the represented timepoint. The likelihood distributions change depending on the prediction horizon. SINDy-det (purple) is less conservative in the bounds than NLL-det (orange) as potential states NLL-det are broader. SINDy-det is more directional and aggressive in predictive actions that generally capture the true trajectory. When using the drift term in SINDY-det to predict where an operator may go, an apt prediction horizon may be suitable to ensure goals are not overshot. This example indicates a reasonable prediction horizon would be less than 2.0s (third column) if a single goal is known. For the distances and teleoperation task in this study, this is a reasonable practical horizon threshold, as operators' can generally reach a velocity of ~1m/s when reaching to grasp an object [39].

Figure 9 gives a showcase of reachability sets for our running example of the operator's motion in the XY plane for the two-goal scenario for the 6 combinations of model fit and goal predictors. The reachability set shows up to a 4s horizon. Each of these methods is shown for different time points (rows and represented by a black dot) with NLL and SINDy modeling processes. The SINDy method is more directed and pointed toward an individual goals (marked by squares, with goal-disc marked as plus sign). The NLL strategies are symmetrical as it tries to cover the potential to move to either goal. The NLL-det method struggles with the introduction of the jumps that force goal changes for short durations. The NLL-det performs a broad model fit to account for changes by expanding confidence bounds significantly to cover the data. In contrast, the SINDy-det and SINDy-NN express high similarity in their distributions demonstrating SINDy's approach to stay more directed toward goals. SINDy-disc has more exploratory behavior heavily reliant on the approximated drift term. This is best seen in the bottom row when comparing SINDy-disc to SINDy-det at t=41s. The SINDy-disc strategy branches out to multiple directions where the SINDy-det stays contained.

Fig. 9: Reachability sets with up to 4s prediction horizons for each strategy combo in the 2 goal scenario in the XY plane. Squares represent goals, green plus sign is current goal-disc and black point is current operator point. The SINDy methods are more directed towards goals due to the drift than NLL that carry more uncertainty. SINDy-disc anticipates wider potential states than SINDy-det as seen in t=41s. NLL-det shows when introducing jumps that force goal changes for short durations, the broad model fit tries to account for changes by expanding confidence bounds significantly to cover them. Meanwhile, the SINDy-det better accounts for sudden jumps in the dynamic estimation and closer resembles an SINDy-NN model estimation.

*2) Quantitative Model Accuracy*

Following the qualitative analysis, a quantitative study was conducted to find if the top predictive actions chosen by these models carry any significant difference. Table III shows the RSS for the predicted trajectory compared to the actual in offline training. A geometric mean and 95% confidence interval on the RSS values were conducted with a 2-way ANOVA on the Strategies (F=3.97, p=0.00136347) and Task Scenarios (F=48.67, p=3.60E-12) to identify differences amongst the groups. The interaction between both led to no significance (F=1.07, p=0.37). A post-hoc Tukey-HSD test was performed on the pairwise differences between the Strategies to identify key differences. The only statistical differences involve NLL-NN compared to NLL-disc (p=0.0186) and all three SINDy approaches (SINDy-NN: p=0.005; SINDy-det: p=0.005; SINDy-disc: p=0.005). No others held significance. The results indicate SINDy predictions are better than the true baseline of NLL-NN.

TABLE III. USER TRAJECTORY 1-STEP PREDICTION SUMMARY

| Strategy | Task Scenario | Offline RSS (m²) | Online RSS (m²) |
|---|---|---|---|
| NLL-det | Single Goal | 0.015 [0.014,0.019] | 0.0012 [0.0011,0.0015] |
| | Two Goal | 0.39 [0.29,0.52] | 0.012 [0.009, 0.016] |
| NLL-disc | Single Goal | 0.016 [0.014,0.017] | 0.0009 [0.0008,0.0012] |
| | Two Goal | 0.39 [0.29,0.51] | 0.01 [0.008,0.0126] |
| NLL-NN | Single Goal | 0.17 [0.13, 0.22] | 0.19 [0.15,0.23] |
| | Two Goal | 2.51 [1.93, 3.25] | 0.29 [0.24,0.36] |
| SINDy-det | Single Goal | 0.019 [0.017,0.022] | 0.0025 [0.0019,0.003] |
| | Two Goal | 0.26 [0.21,0.32] | 0.009[0.007,0.01] |
| SINDy-disc | Single Goal | 0.018 [0.016,0.021] | 0.0025[0.0019,0.003] |
| | Two Goal | 0.26 [0.21,0.0.32] | 0.009 [0.007,0.01] |
| SINDy-NN | Single Goal | 0.019 [0.017,0.022] | 0.0025[0.0019,0.003] |
| | Two Goal | 0.26 [0.21,0.32] | 0.009[0.007,0.01] |

The predictions for online estimation to the actual trajectories to determine the efficacy of the model is studied. Another 2-way ANOVA is performed. The statistical significance between tasks (F=30.94, p=2.8E-8), and strategies (F=369.56, p=2.0E-320) and their interaction (F=19.82, p=1.52E-19) are observed. A post-hoc Tukey HSD test for pairwise differences finds statistical significance for the true baseline NLL-NN with all other strategies regardless of task scenario (all combinations of NLL or SINDy have p<0.0001). No other significance is found for the two-goal scenario. However, in the single goal scenario statistical significance between NLL-disc and all three SINDy methods (SINDy-det: p=0.0299, SINDy-NN: p=0.0299, SINDy-disc: p=0.0248) is observed. While the statistical results show feeding in a continuous goal improves NLL to strikingly low residuals of sub $0.001m^2$, SINDy still performs comparable and better than the true baseline of NLL-NN. From a practical standpoint in observing Table III, the accuracy is nearly millimeter precision, which is inconsequential in our teleoperation scenario due to large actions, and higher noise from measurements due to reliance on an Xbox Kinect camera. Overall, the online and offline results demonstrate SINDy and proposed goal heuristics have improvement over a true baseline of NLL-NN, where the residuals are manageable in the teleoperation scenario.

## IV. Discussion

### A. Jump Identification and Goal Estimation

We demonstrate the power of using the Psychic framework to predict operator motion during multi-goal inference in teleoperation. Specifically, we investigated if jumps are indicators of motion and intent switching. Jumps appear to be present consistently across the trajectories in both scenarios (Table 1). While it is likely operators that are consistent, smooth, and certain of their actions will have reduced jumps, it still is unknown whether this improves stability of their intended tasks. The exploration of identifying key differences between purely diffusive and jumpy operators and the impact on stability of intent inference is left to future work. Specifically, identified jumps may signal an operator's cognitive state of confusion or frustration. Insights in this direction can lead to improvements to the proposed heuristic for the dynamic goal (equation (13)). While SINDy performs comparably in predicting the dynamic goal model in either detection or discovery mode, we see there is a significant difference when goal switching is possible in a scenario. Investigating whether there is a difference between online and offline methods shows inconclusive results. While there is a clear pattern for learned offline models, the online model comparison requires more in-depth exploration in the mechanisms causing the difference. For instance, while the proposed gated dynamics model may be reasonable for offline, an alternative formulation that handles evolving distributions for outliers may be necessary moving forward.

### B. Using Jumps in the Operator Motion Dynamics

We show that using jump-drift-diffusion dynamics paired with SINDy estimation improves intent over a classical regression strategy of NLL. The KM coefficients and SINDy allow for the processes to be decoupled and independently solved. This uniquely gives SINDy a significant advantage in having tighter bounds of possible actions. While SINDy is similar to the NLL method in a controlled setting of a single goal, differences emerge when more goals are introduced. While the best actions may be similar in predictive capabilities, we also demonstrate there is a difference in reachability and prediction horizon sets. Qualitatively, SINDy strategies provide more directionality and cover actions in the event an operator moves drastically. NLL provides potential states that are broad to ensure it captures all data presented. Not only is the model fit improved, but there is the added benefit for an interpretable operator motion dynamics model. While naturally the drift component is the significant piece driving motion, it is beneficial for a future partner to understand an analytical approximation for this function. This is especially beneficial when exploratory states occur (when using the goal-disc mode). This is particularly helpful when used for online approximation. SINDy is readily able to handle lower number of datapoints. We demonstrate there is similar performance gains (against the NLL method) online as there was in offline mode. Overall, this work demonstrates not only the potential, but the benefits of using the SINDy approach. The culmination of this work has led to an available open-access package.

### C. Future Directions to Improve Inference and Control

While this work demonstrates a promising start to intent inferencing with SINDy, there are a few necessary considerations to improve upon in the future. First, designing appropriate basis functions for the task requires domain knowledge, or at least requires trial and error. For instance, designing a basis function that explains drift terms may vary depending on the task. Further, designing a goal trajectory an operator is trying to navigate towards also requires careful curation. While we fit a SINDy model for each individual trajectory, there are developments to build a general inferencing model through multiple trajectories which could be of interest for task specific knowledge [35]. This could be of significant interest where an autonomous agent can decide if a person is acting significantly different from the norm.

Three major considerations are needed to be explored moving forward in jump drift-diffusion modeling for operators. The first is if there is a difference in purely diffusive or jumpy operators. This has implications to determine the cognitive state of an operator and whether intervention in assistance is needed. The second is investigating whether jumps are synchronous or asynchronous across dimensions and their importance. For instance, if operators make sudden changes in a single dimension, it may only indicate a partial change in intent inference, while a synchronized jump may indicate a wholesale change of a goal. The third is determining how to adapt this model to rotational components. While these are long term goals in improving the intent inference, a more immediate consideration is deploying Psychic for control.

Lastly, we leave deploying Psychic for model-based control methods such as MPC and DRL [26] for future work. A main attraction to using Psychic in these methods is the forward rollout of operator actions by using the approximate dynamics model SINDy produces. We envision Psychic would naturally pair with these strategies to improve alignment of the autonomous partners to complete tasks.